\documentclass[letterpaper, 10 pt, conference]{ieeeconf}  

\IEEEoverridecommandlockouts                              

\usepackage{amsmath} 
\usepackage{amssymb}  
\title{\LARGE \bf
Autonomously Navigating a Surgical Tool Inside the Eye by Learning from Demonstration
}

\author{Ji Woong Kim$^{1}$, Changyan He$^{1}$, Muller Urias$^{3}$, Peter Gehlbach$^{3}$,\\ Gregory D. Hager$^{2}$, Iulian Iordachita$^{1}$, Marin Kobilarov$^{1}$
\thanks{}
\thanks{$^{1}$J. Kim, C. He, I. Iordachita, and M. Kobilarov are with Laboratory for Computing + Sensing (LCSR) dept. and $^{2}$GD Hager is with the Computer Science dept. at the Johns Hopkins University, Baltimore, MD 21218 USA {\tt\small \{jkim447, changyanhe, iordachita, marin\}@jhu.edu, hager@cs.jhu.edu}}
\thanks{$^{3}$M. Urias, P. Gehlbach are  with Wilmer Eye Institute at the Johns Hopkins Hospital, Baltimore, MD 21287 USA {\tt\small \{murias1, pgelbach}@jhmi.edu\}.}%
}

\usepackage{graphicx}
\graphicspath{ {images/} }
\usepackage{subfig}
\usepackage{amsmath}
\usepackage{tabularx}
\DeclareMathOperator*{\minimize}{minimize}
\usepackage{makecell}

\begin{document}

\maketitle

\begin{abstract}
A fundamental challenge in retinal surgery is safely navigating a surgical tool to a desired goal position on the retinal surface while avoiding damage to surrounding tissues, a procedure that typically requires tens-of-microns accuracy. In practice, the surgeon relies on depth-estimation skills to localize the tool-tip with respect to the retina in order to perform the tool-navigation task, which can be prone to human error. To alleviate such uncertainty, prior work has introduced ways to assist the surgeon by estimating the tool-tip distance to the retina and providing haptic or auditory feedback. However, automating the tool-navigation task itself remains unsolved and largely unexplored. Such a capability, if reliably automated, could serve as a building block to streamline complex procedures and reduce the chance for tissue damage. Towards this end, we propose to automate the tool-navigation task by learning to mimic expert demonstrations of the task. Specifically, a deep network is trained to imitate expert trajectories toward various locations on the retina based on recorded visual servoing to a given goal specified by the user. The proposed autonomous navigation system is evaluated in simulation and in physical experiments using a silicone eye phantom. We show that the network can reliably navigate a needle surgical tool to various desired locations within 137 $\mu m$ accuracy in physical experiments and 94 $\mu m$  in simulation on average, and generalizes well to unseen situations such as in the presence of auxiliary surgical tools, variable eye backgrounds, and brightness conditions.
\end{abstract}
\section{Introduction}
Retinal surgery is among the most challenging microsurgical endeavors due to its micron scale precision requirements, constrained work-space, and the delicate non-regenerative tissue of the retina. During the surgery, one of the most challenging tasks is the spatial estimation of the surgical tool location with respect to the retina in order to precisely move its tip to a desired location on the retina. For example, when performing retinal-peeling or vein cannulation, the surgeon must rely on intuitive depth-estimation skills to navigate toward a targeted location on the retina, while ensuring that the tool-tip contacts the retina precisely at the desired location. Such maneuvers introduce high risk because the surgical tools are sharp and the slightest misjudgement can damage the surrounding tissues, which could lead to serious complications.
\begin{figure}[h]
        \centering
        \includegraphics[width=0.43\textwidth]{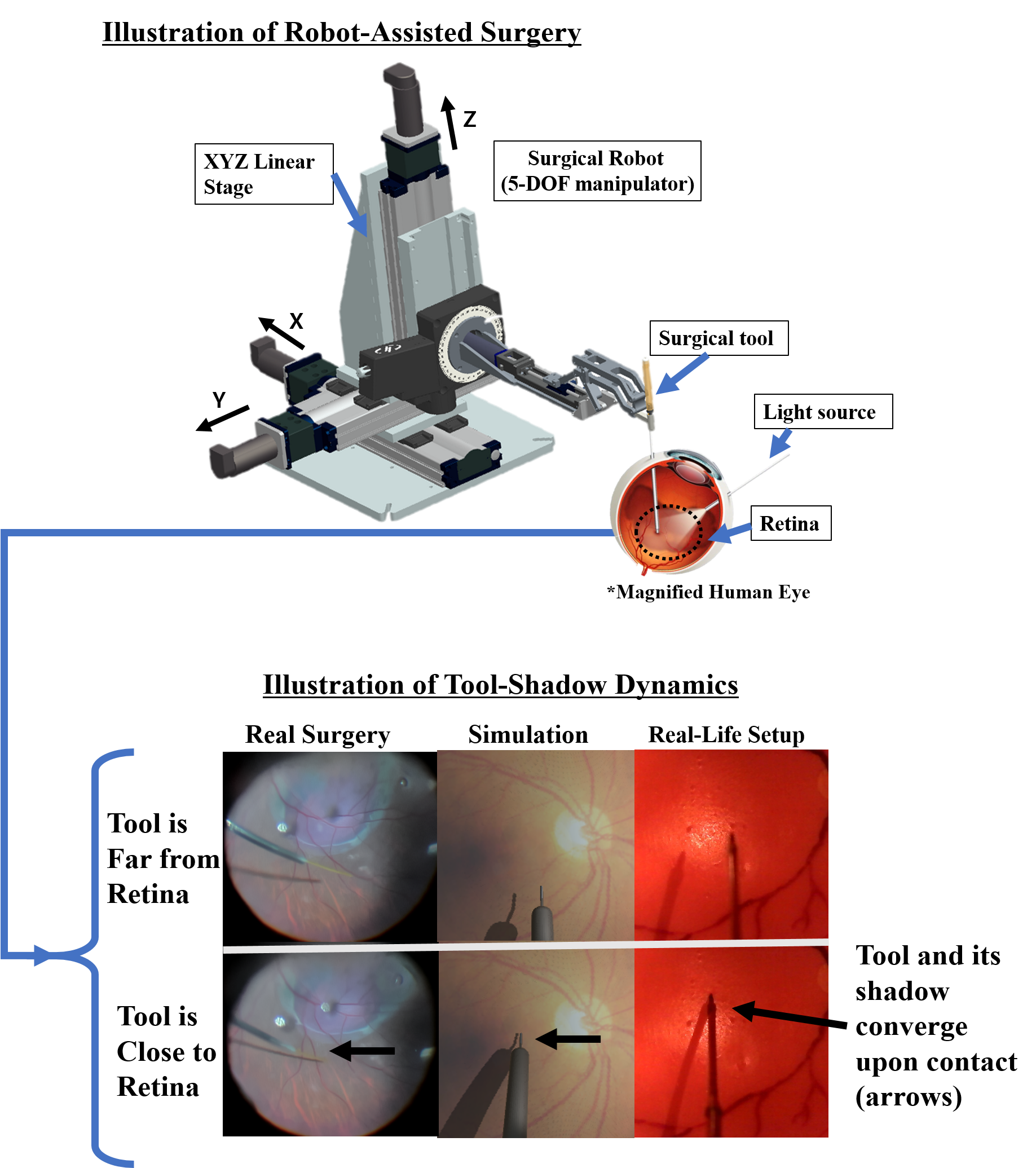}
        \caption{ \small (Top) During robot-assisted retinal surgery, a light source projects a shadow on the retina which can be used as cues to estimate proximity between the tool-tip and the retina. (Bottom) Demonstration of tool-shadow dynamics; as the surgical tool approaches close to the retina, the tool and its shadow converge (compare top row to bottom row), which can be used as cues to train a network how to navigate inside the eye.}
        \label{Fig-Intro}
\end{figure}

\label{sec:result}
\begin{figure*}[h]
        \centering
        \includegraphics[width=0.8\textwidth]{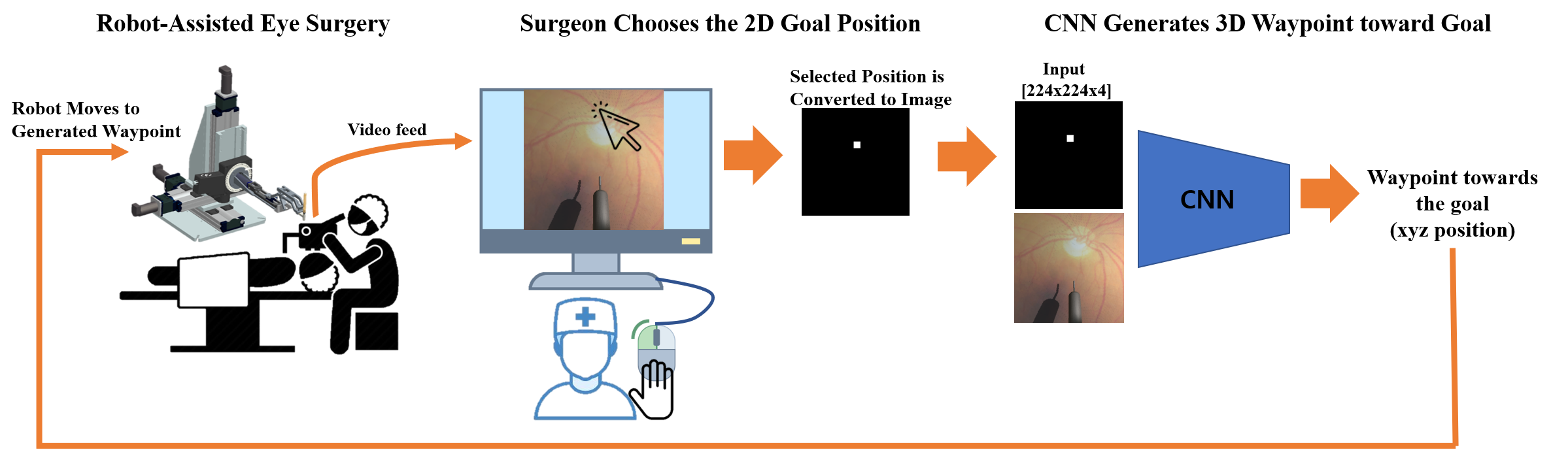}
        \caption{\small System setup: a surgeon chooses the goal location in 2D and the network generates a 3D waypoint that navigates the surgical tool toward the selected location.}\vspace{-20pt}
        \label{fig:system_setup}
\end{figure*}
To alleviate the difficulty of the tool-navigation task, we propose to automate it by learning the closed-loop visual servoing process employed by surgeons, i.e. mapping from visual input (video) to euclidean position control commands to actuate the robot. 
Specifically, we train a deep network to imitate expert trajectories toward various locations on the retina based on many demonstrations of the tool-navigation task. The input to the network are the monocular top-down view of the surgery through a microscope  and user-input defining the 2D goal location to be reached. 
The advantage of this method is that the user only specifies the goal in 2D, e.g. as simple as clicking the desired location using a mouse (Fig. \ref{fig:system_setup}), and the network outputs a 3D waypoint toward the target location on the retina. Since estimating depth is the challenging task for humans, the network takes the burden of extrapolating how to navigate along the depth dimension based on its training experience. Learning such simple tool-navigation maneuver is fundamental in automating surgery, since it is the primitive action performed in any surgical procedures. 

We note that our approach is grounded in the hypothesis that the tool-navigation task may be automated primarily using vision. In fact, surgeons rely on their vision to localize objects and estimate their spatial relationship to navigate the surgical tool. Furthermore, the surgical scene captures a distinct tool-shadow dynamics which can be useful for recognizing proximity between the tool-tip and the retina. Specifically, the tool and its shadow converge upon approaching the retina (Fig. \ref{Fig-Intro}), which can be as cues to train the network. In addition, while a complete setup can include stereo vision, in this work we rely on a single camera alone for simplicity. We also utilize a force-sensing modality to detect contact with retina, such that the surgical tool can be stopped upon contact. 

The system performance is validated experimentally using both an artificial eye-phantom as well as in simulation employing the Unity3D (Unity Technologies) environment \cite{unity_ml_agents}. The main objective is to assess the quality of surgical tool navigation to desired locations on the retina. To achieve this, we employ a batch of benchmark tasks where various positions on the retina are targeted in a grid-like fashion (Fig. \ref{fig:real_life_test_condition}). For simplicity, we keep the eye position and tool-orientation fixed during the experiments. While this is not a realistic assumption in practice, since the eye could involuntarily move during procedures, our approach can easily extend to the more general setting of different eye rotations and tool orientations through additional training. To test the robustness of our network, we also perform the benchmark task in the presence of unseen distractions in the visual input, such as a light-pipe (used for illuminating the surgery scene) and forceps (used in retinal-peeling) which are commonly used surgical tools. On average, we report that the network achieves 137 $\mu m$ accuracy in various unseen scenarios in physical experiments using a silicone phantom, and 94 $\mu m$ accuracy in simulation on average.  Lastly, we propose a change to the baseline network resulting in marked improvement in its performance, specifically by training the network using future images along with waypoints as labeled outputs, which turns out to be a richer representation useful for control. We demonstrate that learning such auxiliary task improves the performance on the tool-navigation task.
\par


\label{sec:result}
\section{Related Work}
\begin{figure*}[h]
        \centering
        \includegraphics[width=0.70\textwidth]{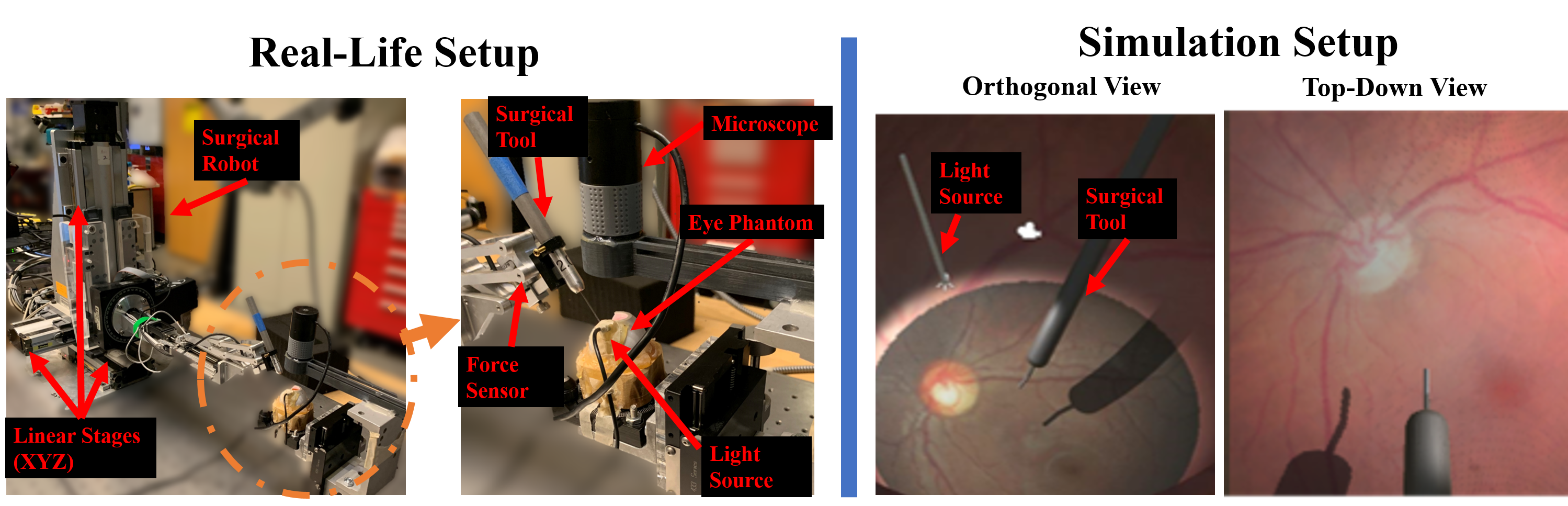}
        \caption{\small (Left) Real-life experimental setup using an eye phantom. (Right) Experimental setup in simulation}\vspace{-20pt}
        \label{fig:exp_setup}
\end{figure*}
\subsection{Retinal Surgery}
Past works in computer-assisted retinal surgery have focused on state estimation or detection systems to assist surgeons with more information about the surgery. For example, image segmentation can be applied to estimate tool-tip and shadow-tip to model proximity when the tips approach close by a predefined threshold pixel-distance~\cite{japan_shadow}. In addition, stereo vision can be employed to estimate the depth of the tool and the retina respectively to create a proximity detection system ~\cite{6212340}. More recently, optical coherence tomography (OCT) was utilized to sense depth between the tool-tip and the retina~\cite{Ourak2019, leuven_OCT}. While these methods do not address automation, they are relevant in the sense that if one could estimate distance from tool-tip to goal, then autonomous navigation may be achieved by interpolating between the two points. Although stereo-vision method may be one alternative, it is not expected to work reliably in a clinical setting due the unknown distortion caused by the patient's lens and out-of-focus images, which may cause depth reconstruction error. OCT is also a promising alternative, however, OCT measures distance locally and is challenging to attach on tool-tips with non-cylindrical geometries such as forceps, which are commonly used surgical tools. On the other hand, the learning based approach proposed in this paper can be trained to be robust to visual distortions and navigate complex-shaped surgical tools, given that appropriate training data is available. This is possible since deep networks succeed in the task by imitating the expert, who always succeeds in the task.



\par
\par

\subsection{Learning}
Various works have shown the effectiveness of deep learning in sensorimotor control such as playing computer games \cite{atari_games, learn_to_act} or navigating in complex environments \cite{nav_1, drive_2, drive_3, MIT_autonomous_driving, conditional_learning, chauffeur_net}. In particular, the approach employed in our work borrows from the architecture proposed in~\cite{conditional_learning}, where a network is trained to drive a vehicle based on user's high level commands such as "go left" or "go right" at an intersection. Similarly, \cite{chauffeur_net, MIT_autonomous_driving} employ topographical maps to communicate the desired route to a destination selected by a user to drive a vehicle. In our work, we also communicate the goal position as a topographical representation to navigate the surgical tool (Fig. \ref{fig:system_setup}). Furthermore, several prior works employ the idea of learning auxiliary tasks to improve accuracy, such as predicting high-dimensional future image conditioned on input (e.g. goal or action) ~\cite{video_prediction, facebook_vid2game, visual_task_planning, deep_visual_foresight, cGAN}, and learning auxiliary tasks for improved sensorimotor control~\cite{nav_2, chauffeur_net}.

\section{Problem Formulation}
We consider the task of autonomously navigating a surgical tool to a desired location using a monocular surgical image and topographical 2D goal-position specified by the user as inputs (Fig. \ref{fig:system_setup}). We formulate the problem as a goal-conditioned imitation learning scenario, where the network is trained to map observations and associated goals to actions performed by the expert. The goal-conditioned formulation is necessary to enable user-control of the network at test time (i.e. navigate the surgical tool to a desired location). Given a dataset of expert demonstrations, $D=\{(o_i, g_i), a_i\}_{i=1}^{N}$, where $o_i$, $g_i$ and $a_i$ denote observation, goal, and action, respectively, the objective is to construct a function approximator $a=F(o;\theta)$ with parameters \(\theta\), that maps observation-goal pairs to actions performed by the expert. The objective function can then be expressed as the following: 
\begin{gather}
\minimize_{\theta} \sum_{i=1}^{N} L(a_i, F(o_i, g_i; \theta)),
\end{gather}
where $L$ is a given loss function.

In our case, we choose the observation to be an monocular image $o\in \mathcal I$ of the surgical scene viewed from top-down,  the action $a\in\mathcal W\subset \mathbb{R}^3$ to be the 3D euclidean coordinates of a point in the surgical workspace $\mathcal W$ or a waypoint, and the goal input to be $g_i=(x_i,y_i)\in\mathbb{R}^2$ which specifies the final desired projected 2D position on the retinal surface. Further details on how the expert dataset is collected and network is trained are given next.

\section{Method}

\begin{figure*}[h]
        \centering
        \includegraphics[width=0.75\textwidth]{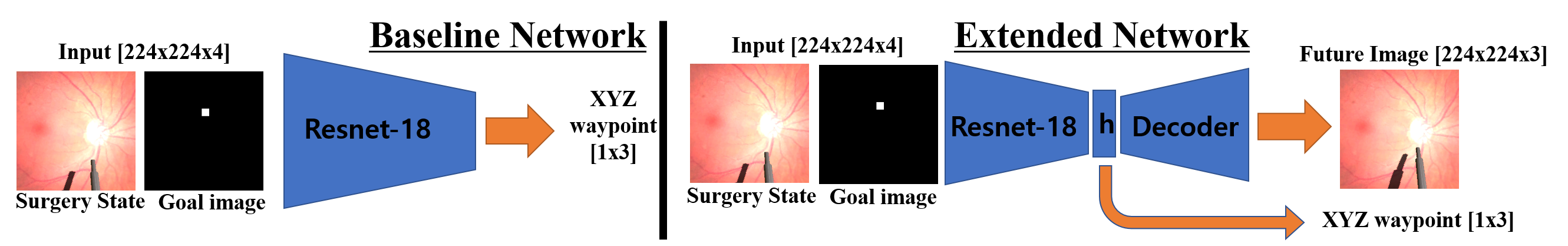}
        \caption{(Left) Baseline network (Right) Extended network}
        \label{fig:network_architecture}
\end{figure*}

\begin{figure}[h]
        \centering
        \includegraphics[width=0.40\textwidth]{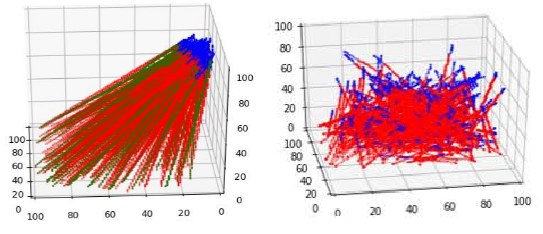}
        \caption{\small (Left) training trajectories with small initialization region; blue tips indicate starting position of the trajectory (Right) Training trajectories with large initialization space as data augmentation.}\vspace{-15pt}
        \label{fig:data_augmentation}
\end{figure}
\subsection{Eye Phantom Experimental Setup}
Our experimental setup consists of the robot, a surgical needle, and a microscope that records top-down view of the surgery as shown in Fig.~\ref{fig:exp_setup}. For our robot platform, we used the Steady Hand Eye Robot (SHER), which is a surgical robot built specifically for eye surgery applications \cite{Steadyhand2}. The surgical needle is attached at the end-effector with thickness 500$\mu m$ in diameter and its small tip measuring 300 $\mu m$ in diameter (Fig. \ref{fig:real_life_test_condition}). The artificial eye phantom (i.e. a rubber  eye model) is 25.4mm in diameter, slightly larger than a human eye which ranges 20 - 22.4mm~\cite{eye_diameters}. To collect data in our experiments, we control the robot using motors attached on the robot joints. We record the images from the microscope and the tool-tip position based on the robot motor encoders.
\subsection{Simulation Setup}
\begin{figure*}[h]
        \centering
        \includegraphics[width=0.75\textwidth]{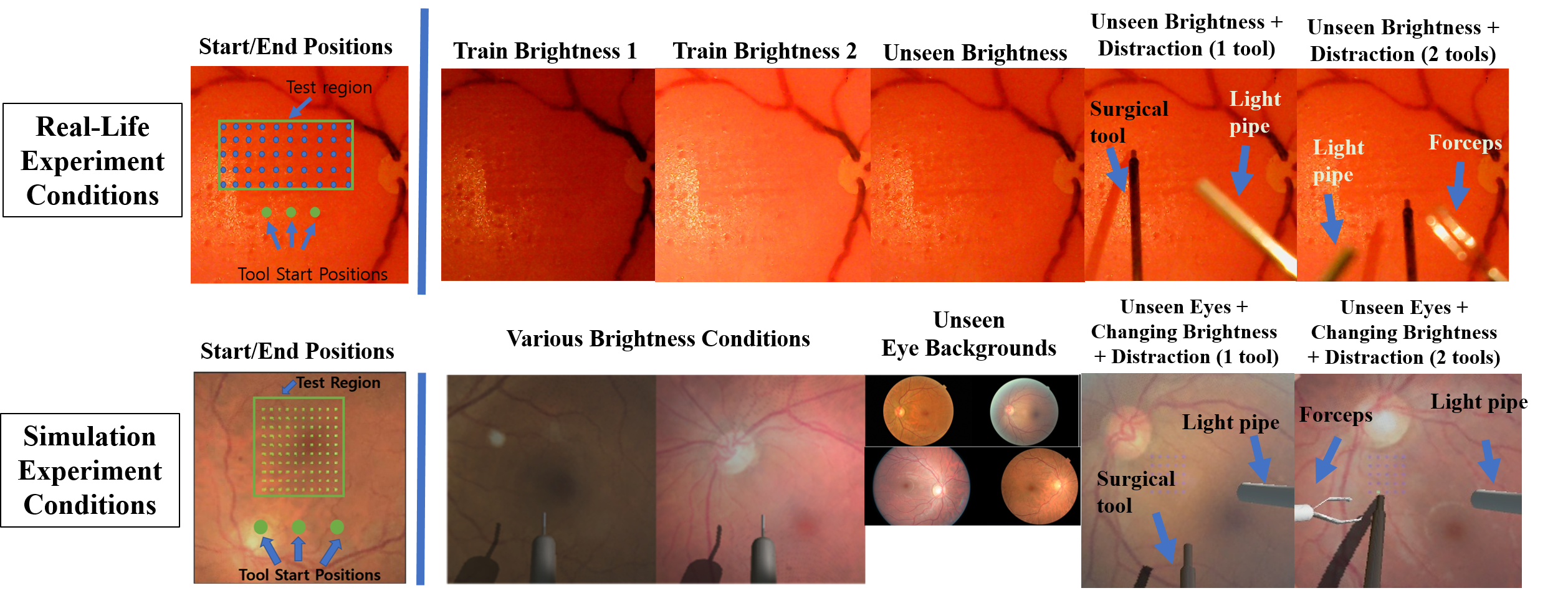}
        \caption{\small (Top) Example test conditions in real-life physical experiment and (Bottom) in simulation.}\vspace{-15pt}
        \label{fig:real_life_test_condition}
\end{figure*}

In simulation, we used Unity3D software to replicate similar experiment scenarios as the physical experiment as shown in Fig.~\ref{fig:exp_setup}. For sense of scale, the thick part of the tool shaft measures 500$\mu m$ and the tool-tip measures 300$\mu m$  (Fig.~\ref{fig:real_life_test_condition}). We perform domain randomization to change the eye background texture and the lighting condition. Specifically, we created 15 different eyes, each varying in dimension at 20.4mm, 21.2mm, and 22.4mm. These measurements reflect the minimum, medium, and maximum dimension of human eye sizes \cite{eye_diameters, eye_thickness}, and 5 eyes were created for each dimension. The texture of the eyes were obtained from \cite{kaggle_retinopathy}.  Domain randomization was used to help the network generalize-well to changing brightness conditions, size of the eye, and unseen eye background textures. three eyes from each size were used in training, and the remaining two eyes from each size were used for testing.

\subsection{Data Collection}\label{sec:data}

For real-life experiments, we collected 2000 trajectories in low and high brightness settings. In simulation, we collected 2500 trajectories under a wide range of brightness conditions, while various eyes with different size and backgrounds were randomly replaced. The procedure for data collection were as follows: we initialized the tool at a random position, then navigated towards a randomly selected position below the eye in a straight-line trajectory. We use straight-line trajectories as a way to generate predictable and simple ground-truth expert data. When collision was detected between the tool and the eye phantom using a force sensor at the end-effector (Fig. \ref{fig:exp_setup}), we logged the images and the tool-tip positions of the trajectory. For simplicity, we kept the orientation of the tool fixed and only moved the XYZ-position of the tool. The position of the eye remained fixed as well.

To synthesize the goal for each trajectory, we used the last XY tool-tip position of the executed trajectory and plotted it as a white square with dark background as illustrated in Fig.~\ref{fig:system_setup}. Effectively, we changed the 2-dimensional coordinate representation of the goal to an image representation. This design choice can be useful in an application where a surgeon may simply use a mouse to click the goal location in the visual-feed of the surgery, and the network will navigate the surgical tool to the exact location of the white square corresponding to the surgical scene (Fig.\ref{fig:system_setup}). 

We note that our method for synthesizing goal images do not guarantee precise spatial correspondence between the tool-tip position and the goal image. This is due to perspective projection, where objects further away from the camera are subject to shift towards the vanishing point, but the plotted goal position, which is obtained from robot kinematics, is not subject to perspective projection. One way to guarantee spatial correspondence is to manually annotate the tool-tip positions to create the goal images; however, we leave this for future work. Our objective is to assess how consistently the network can generate desired trajectories given a particular goal image. Thus, similar consistency in performance is expected given manually-annotated goal images with precise spatial correspondences.

\subsection{Network Details and Training}
The input to the network are the current image of the surgery (224x224x3) and the goal image (224 x 224 x 1) stacked along the channel dimension, yielding a combined dimension of (224 x 224 x 4). The output of the network is a XYZ-waypoint in the surgical workspace (three-dimensional vector) which the network must travel in order to reach closer to the target location on the retina. Specifically, for a single trajectory consisting of $n$ frames $I_1, ... I_n \in \mathcal I$, $n$ tool-tip positions $p_1,\dots, p_n \in \mathbb{R}^3$, and the goal image coordinate $g \in \mathbb{R}^2$ specific to this trajectory, a single sample is expressed as $(\text{input}, \text{output})=\left((I_t, g), p_{t+d}\right)$, for $t=1,\dots,n\!-\!d$, where $d$ is a parameter denoting the \emph{look-ahead} of the commanded action, which is used as a feed-forward reference signal to the robot. We chose \(d=8\), which is equivalent to approximately 70$\mu m$ apart between learned waypoints to ensure that the network moved the surgical tool by a noticeable distance every control cycle. The complete data set $D$ is constructed using multiple such trajectories and their corresponding samples. Internally, the network maps the goal $g$ into an image $I_g\in\mathcal I$ which is concatenated with the actual camera image $I_i$ to form the complete network input. 

We experimented with two architectures, a baseline network that predicts an XYZ waypoint and another network that predicts an XYZ waypoint plus the future image as shown in Fig. \ref{fig:network_architecture}, which we refer to as extended network. The extended network was enforced to learn an auxiliary task which helped with the main objective of predicting accurate waypoints. For the extended network, a single training sample is $(\text{input}, \text{output}) = ((I_t, g), (p_{t+d}, I_{t+d}))$. In the following, we discuss each network in greater detail. 

\begin{table}[h]
\caption{\label{tab:real_life_table} Eye Phantom Experiment Results}
\setlength\tabcolsep{3pt}
\begin{tabular}{c c c}
\hline
Test Condition                         & \thead{Baseline Network \\ Error (mm)} & \thead{ Extended Network \\ Error (mm)} \\
\hline

Train Low Brt.1                        & 0.134                       & 0.139                       \\

Train High Brt. 2                        & 0.092                       & 0.108                       \\

Unseen Brt.                         & 0.177                       & 0.127                       \\

\thead{Unseen Brt.+Distr. (1 tool)}  & 0.165                       & 0.146                       \\

\thead{Unseen Brt.+Distr. (2 tools)}& 0.155                       & 0.137                       \\

\thead{"Unseen" Avg. (above 3 rows)}   & \textbf{0.166}                       & \textbf{0.137}                       \\


\hline
\vspace{-5pt}
\end{tabular}
\end{table}

\vspace{-20pt}

\begin{center}
\begin{table}[h]
\centering
\caption{\label{tab:real_life_train_results}Eye Phantom Training Results}
\vspace{-5pt}
\begin{tabular}{c c c}
\hline
Axes                        & \thead{Baseline Val. \\ Acc. (\%)} & \thead{ Extended Network \\ Val. Acc. (\%)} \\
\hline

X                      & 82.0                       & 82.8                       \\

Y                        & 76.0                       & 76.7                       \\

Z (Depth)                         & 60.8                       & 61.9                       \\

XYZ Total Sum                & \textbf{218.8}                             &\textbf{221.3} \\
\hline

\end{tabular}
\end{table}
\end{center}
\vspace{-20pt}

\subsubsection{Baseline Network}
We use Resnet-18 \cite{resnet}, which takes high dimensional image (224 x 224 x 4) as input to 512-dim feature vectors. To learn the waypoints or the action output, we discretize the continuous X, Y, and Z coordinate representation into 100 steps. Specifically, we add a fully-connected layer outputting 300 neurons on top of the 512-feature vectors, where each 100 neurons is a discretized representation of the continuous X, Y, and Z coordinates of the euclidean surgical workspace respectively. The network was trained using cross-entropy loss with Adam optimizer ~\cite{adam_optimizer} with a learning rate of 0.0003 and batch size of 170. The loss function is defined as
\begin{align}
    L(b, \hat p) = \sum_{j\in \{x,y,z\}} \sum_{c=1}^{M_j} -b_{j,c} \log(\hat p_{j,c}),
\end{align}
where $b_{j,c}$ are binary indicators for the true class label $c$, and $\hat p_{j,c}$ are the predicted probability that the coordinate $j$ is of class $c$. The cost combines the errors for all three dimensions $j\in \{x,y,z\}$. As specified above, we employed $M_x=M_y=M_z=100$ discrete bins.

\subsubsection{Baseline + Predicting Future Image (Extended Network)}
The extended architecture aims to achieve the baseline task and additionally predict future images. The architecture is shown in Fig.~\ref{fig:network_architecture}. On top of the Resnet-18 architecture, a decoder network with skip-connections is added. The waypoints were trained using cross-entropy loss similar to the baseline network and the future-image prediction was trained using RMSE function. The network is trained using Adam optimizer with a learning rate of 0.0003 and batch size of 120. The combined loss function is given as
\begin{align}
    \!\!\!L((b, I),(\hat p, \hat I))\!=\!\!\!\sum_{j\in \{x,y,z\}} \sum_{c=1}^{M_j} -b_{j,c} \log(\hat p_{j,c})\!+\!(I\!-\!\hat I)^{2},
\end{align}
where $\hat I$ denotes the future-image prediction by the network and $I$ denotes the label for the future image. To balance the loss functions, drop-out approach was used where we performed back-propagation 70\% of the time for the future-image loss term.

\subsection{Data Augmentation}
For robust learning, we utilized data augmentation such as random drop-out of pixels, Gaussian noise, and random brightness, contrast, and saturation. We also expanded the initialization space so that the network could reach the same target location from various initial positions as shown in Fig. \ref{fig:data_augmentation}. This effectively enabled the network to recover from mistakes when it deviated from its hero path \cite{dagger}. These techniques were crucial to generalizing well to unseen environments.

\section{Results and Discussion}
\subsection{Physical Experiment Results}

To assess the accuracy of our networks, we performed benchmark experiments where the baseline and the extended network visited 50 predefined locations in the training region in grid-like fashion (5 x 10), starting from three different initial locations  as shown in Fig. \ref{fig:real_life_test_condition}. The objective of such experiment was to test how accurately the network could navigate to various targeted locations, given various goal inputs. We tested each network in the  familiar and unseen environments as listed in Table \ref{tab:real_life_table} and Fig. \ref{fig:real_life_test_condition}. For experiments with tool-distractions, we only tested from the right-most initial position out of the three, since a human had to hold the tools throughout the long experiments. The light-pipe and forceps were dynamically maneuvered to follow the tool-tip. Both tools occasionally occluded the surgical tool and its shadow. 

\begin{figure}[h]
        \centering
        \includegraphics[width=0.5\textwidth]{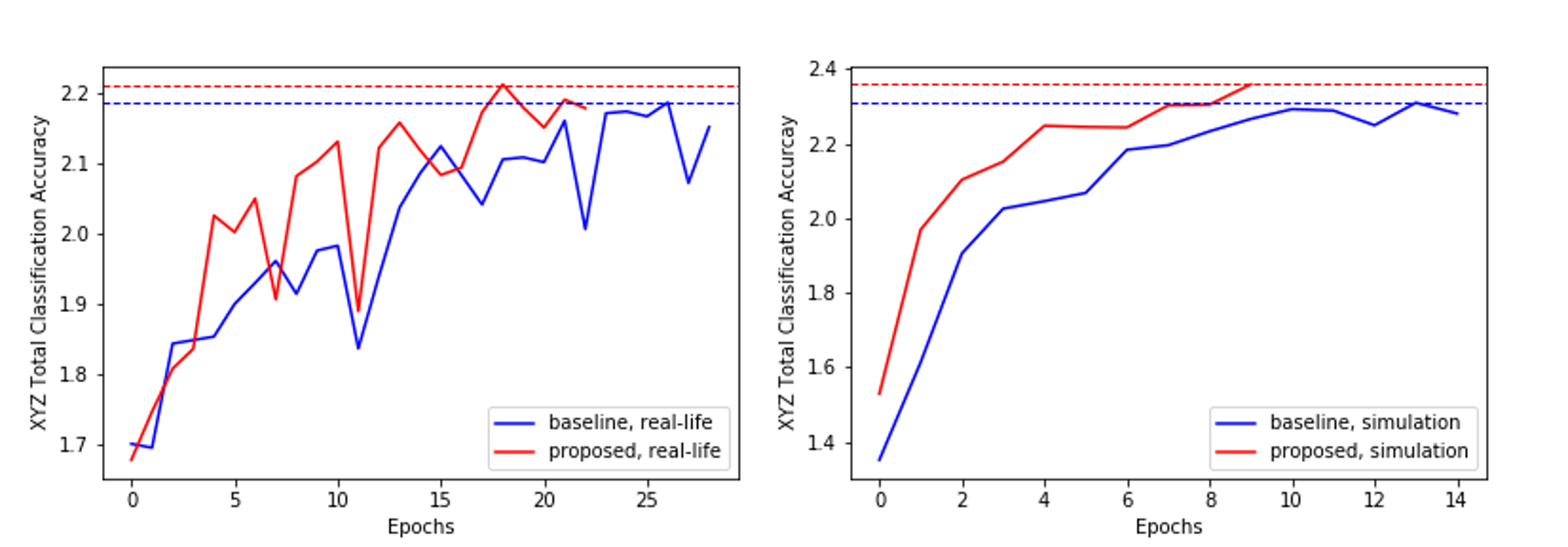}
        \caption{\small (Left) Waypoint prediction accuracies on the "real-life" validation dataset and in (Right) simulation. Y-axis is the sum of the classification accuracies for xyz axes (i.e. maximum possible is 3.0). Dashed lines mark the maximum accuracy achieved.}\vspace{-5pt}
        \label{fig:real_life_sim_plot}
        \vspace{-5pt}
\end{figure}

\begin{figure*}[h]
        \centering
        \includegraphics[width=0.75\textwidth]{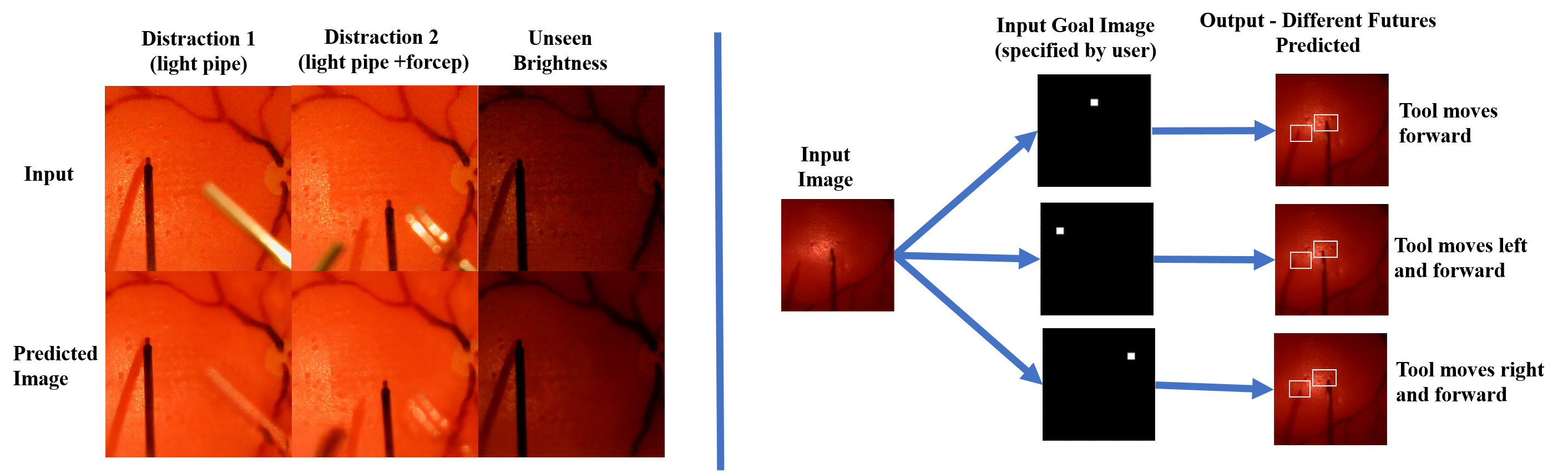}
        \caption{\small (Left) Future prediction by the extended network under various unseen conditions (Right) Different futures predicted by changing the goal input (rectangle frames are fixed, added to clarify shifted positions of the tool)}
        \label{fig:future_predictions}
\end{figure*}

Our experimental results are summarized in Table \ref{tab:real_life_table} and the executed trajectories are shown in Fig. \ref{fig:real_life_traj}. The table contains numeric XY-error values in reaching the goal position under various test conditions. Since the eye position is fixed during training and experimental validation, the error can be calculated by comparing the input goal-image coordinate \((x,y)\) against the final landing position of the surgical tool \((x^{'},y^{'})\) after the trajectory execution is complete (e.g. when force is detected using the force sensor). Thus, the error reported in Table \ref{tab:real_life_table} is calculated using the formula \(\sqrt{(x-x^{'})^2+(y-y^{'})^2}\). The accuracy reported in Table \ref{tab:real_life_train_results} are the classification accuracies achieved on the validation dataset offline, not the online benchmark experiments. In Table \ref{tab:real_life_train_results}, we are able to report errors in the z-axes (depth) because we have ground-truth xyz-values of the full trajectory from the previously collected dataset. 


Our results show that the both baseline and extended network generalizes well to unseen scenarios, achieving 166$\mu m$ and 137$\mu m$ in error, even in the presence of unseen brightness conditions and unseen surgical tools significantly occluding the scene. The extended network also performed marginally better than the baseline network. This result is expected given the higher accuracy achieved by the extended network in the validation dataset, achieving 2.5\% higher accuracy than the baseline (Table \ref{tab:real_life_train_results}). Also, as shown in Fig. \ref{fig:real_life_sim_plot}, the extended network trains faster and is more data-efficient than baseline network, achieving best classification accuracy on the 18th epoch versus 26th epoch by the baseline network. In addition to improving the baseline network performance, the extended network is able to predict clear future images. The extended network can imagine different futures depending on various goal inputs (e.g. move the tool forward, left, right), recognize the surgical tool as a dynamic object apart from the static background, and also reliably reconstruct surgical objects it has never seen during training (Fig. \ref{fig:future_predictions}).

 \subsection{Simulation Experiment Results}
Similar to real-life experiments, we performed benchmark tests where each baseline and extended network visited 100 predefined locations in the training region in grid-like fashion (10 x 10), starting from three different initial locations as shown in Fig. \ref{fig:real_life_test_condition}. We tested each network in the following conditions as listed in Table \ref{tab:sim_table} and shown in Fig.\ref{fig:real_life_test_condition}. For experiment with tool-distractions, we only tested from the middle initial position out of the three. Both the light-pipe and forceps were moved randomly to imitate hand-tremor, and both tools occasionally occluded the surgical tool and its shadow. 

\begin{figure}[h]
        \centering
        \includegraphics[width=0.45\textwidth]{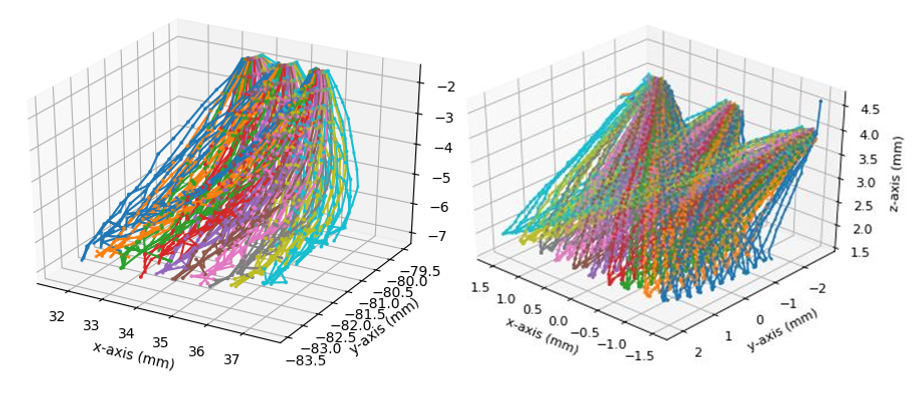}
        \caption{(Left) Trajectories executed in real-life in unseen brightness condition (Right) trajectories executed in real-life in changing brightness condition + unseen eyes}
        \label{fig:real_life_traj}
\end{figure}

The simulation results are summarized in Table \ref{tab:sim_table} and the executed trajectories are shown in Fig. \ref{fig:real_life_traj}. The errors shown in Table \ref{tab:sim_table} are calculated using the same formula mentioned in the real-life experiment results, specifically using the formula \(\sqrt{(x-x^{'})^2+(y-y^{'})^2}\) where \((x,y)\) denotes input goal-image coordinate and  \((x^{'},y^{'})\) denotes the final landing position of the surgical tool after trajectory execution. Similarly, Table \ref{tab:sim_table_train_results} shows network results on the validation dataset. Our results show that both baseline and extended networks achieve good performance and can generalize robustly to unseen scenarios, even in the presence of unseen eye backgrounds and unseen surgical tools occluding the scene. Similar to real-life experiments, the extended network also performed marginally better than the baseline network. This result is expected since the extended network achieved 4.9\% higher accuracy than the baseline network in the validation dataset (Table \ref{tab:sim_table_train_results}). Similar to real-life experiments, the extended network is also more data-efficient than the baseline network, achieving maximum accuracy at 13th epoch versus 9th epoch by the baseline network (Fig.\ref{fig:real_life_sim_plot}) and achieving significantly higher maximum validation accuracy.

\vspace{-20pt}
\begin{table}[h]
\caption{\label{tab:sim_table}Simulation Experiment Results}
\vspace{-5pt}
\setlength\tabcolsep{3pt}
\begin{tabular}{c c c}
\hline
Testing Condition                      & \thead{Baseline Network \\ Error (mm)} & \thead{ Extended Network \\ Error (mm)} \\
\hline
Train                                   & 0.107                       & 0.098                       \\
Unseen Eyes                             & 0.102                       & 0.096                       \\

\thead{Unseen Brt. + Distr. (1 tool)}      & 0.140                       & 0.100                       \\

\thead{Unseen Brt. + Distr. (2 tools)}   & 0.169                       & 0.087                       \\

\thead{"Unseen" Avg. (above 3 rows)} & \textbf{0.137}                       & \textbf{0.094}                       \\

\hline
\vspace{-20pt}
\end{tabular}
\end{table}
\vspace{-5pt}

\begin{center}
\vspace{-10pt}
\begin{table}[h]
\centering
\caption{\label{tab:sim_table_train_results}Simulation Training Results}
\begin{tabular}{c c c}
\hline
Axes                        & \thead{Baseline Val. \\ Acc. (\%)} & \thead{ Extended Network \\ Val. Acc. (\%)} \\
\hline

X                      & 78.9                       & 81.4                       \\

Y                        & 84.8                       & 84.0                       \\

Z (Depth)                         & 67.3                       & 70.6                       \\

XYZ Total Sum                & \textbf{231.0}                             &\textbf{235.9} \\
\hline

\end{tabular}
\vspace{-10pt}
\end{table}
\end{center}

\vspace{-25pt}
\section{Conclusions}
In this work, we show that deep networks can reliably navigate a surgical tool to various desired locations within 137 $\mu m$ accuracy in physical experiments and 94 $\mu m$  in simulation on average. In future work, we hope to consider more realistic scenario including a physical scelera constraint, eye movement, and also using porcine eyes.
\vspace{-5pt}
\section{Acknowledgements}
This work was supported by U.S. NIH grant 1R01EB023943-01. 




\bibliographystyle{IEEEtran} 
\bibliography{bib} 

\end{document}